# Specifying Logic Programs in Controlled Natural Language

Norbert E. Fuchs, Rolf Schwitter
Department of Computer Science, University of Zurich
{fuchs, schwitter}@ifi.unizh.ch

**Abstract**
Writing specifications for computer programs is not easy since one has to take into account the disparate conceptual worlds of the application domain and of software development. To bridge this conceptual gap we propose controlled natural language as a declarative and application-specific specification language. Controlled natural language is a subset of natural language that can be accurately and efficiently processed by a computer, but is expressive enough to allow natural usage by non-specialists. Specifications in controlled natural language are automatically translated into Prolog clauses, hence become formal and executable. The translation uses a definite clause grammar (DCG) enhanced by feature structures. Inter-text references of the specification, e.g. anaphora, are resolved with the help of discourse representation theory (DRT). The generated Prolog clauses are added to a knowledge base. We have implemented a prototypical specification system that successfully processes the specification of a simple automated teller machine.

## 1 Declarative Specifications

The derivation of formal software specifications from informal requirements is not easy and cannot be formalised. However, the derivation process can be made easier by the deliberate choice of a specification language that allows users to express concepts of an application domain concisely and directly, and to convince them of the adequacy of the specification without undue difficulty.

Though Prolog has been recommended as a general high-level specification language, and has often been used as such, application-specific specification languages seem to be a better choice since they allow users to express the concepts of the application domain directly, and still can be mapped to Prolog [Sterling 92]. By making "true statements about the intended domain of discourse" [Kramer & Mylopoulos 92] and "expressing basic concepts directly, without encoding, taking the objects of the language as abstract entities" [Börger & Rosenzweig 94], application-specific specification languages are – in the true sense of the word – declarative, and have all the advantages of declarative programming [Lloyd 94].

In a previous phase of our project we have already shown that graphical and textual views of logic programs can be considered as application-specific specification languages [Fuchs & Fromherz 94]. Graphical views include transition networks for finite state automata and window-oriented user-interfaces, while textual views comprise formal specification languages.

Each view has an associated editor that allows to compose specifications from predefined and reusable elements of a repository. Furthermore, there is an automatic bi-directional mapping between a program and its views.

Both these features have important consequences.

- With the help of the view editors we can compose programs by means of application-specific concepts.
- The mapping of a view to a program in a logic language assigns a formal semantics to the view. Thus, though views give the impression of being informal and have no intrinsic meaning, they are in fact formal and have the semantics of their associated program.

- The executability of the logic program and the semantics-preserving bi-directional mapping between a program and its views enable us to simulate the execution of the program on the level of the views. Thus validation and prototyping in concepts close to the application domain become possible.
- By providing semantically equivalent representations, we can reduce the gap between the different conceptual worlds of the application domain specialist and of the software developer. In addition, the dual-faced informal/formal appearance of the views provides an effective solution for the critical transition from informal to formal representations.

Altogether, the characteristics of the views let us call them specifications of the program. Furthermore, since the views are semantically equivalent to the program, they can even be considered as executable specifications.

In the following, we describe an approach using controlled natural language – a subset of natural language characterised by a restricted grammar and an application-specific vocabulary – as a further view of a logic program. Users compose specifications for logic programs in controlled natural language that are automatically translated into Prolog clauses. As pointed out above, this translation makes seemingly informal specifications in controlled natural language formal, assigns them a semantics, and gives them the combined advantages of informality and formality. The generated Prolog knowledge base can be queried and executed. Its clauses can also be paraphrased in controlled natural language. However – contrary to the system described in [Fuchs & Fromherz 94] – the exact original cannot be reproduced since the controlled natural language system uses a finely-grained lexicon instead of a repository of larger chunks of interrelated information.

We have implemented a specification system offering the following functionality. The user enters interactively specification text in controlled natural language that is parsed by a DCG enhanced by feature structures, analysed for discourse references, and translated into Prolog clauses that are added to a knowledge base. The user can ask questions that are processed as Prolog queries and answered with the help of the knowledge base. Our main specification example is a simple automated teller machine.

Section of this paper 2 describes the motivation for controlled natural language and delineates its syntactical constructs. In section 3 we briefly introduce phrase structure grammars and GULP, a linearised notation for feature structures. Section 4 summarises discourse representation theory and gives examples of simple and complex discourse representation structures. Furthermore, we show how eventualities – events and states – can be represented by discourse representation structures in a natural way. In section 5 we give an overview of the prototypical specification system that we have been implementing. Section 6 shows the translation of an example sentence in controlled natural language into a discourse representation structure, and then into Prolog. This section also addresses the paraphrasing of specifications, query answering, and the execution of a specification. In section 7 we conclude and outline future research.

## 2 Controlled Natural Language

A software specification is a statement of the services a software system is expected to provide to its users. It should be written in a concise way that is understandable by all potential users of the system [Sommerville 92]. Strangely enough, this goal is hard to achieve if specifications are expressed in full natural language. Natural language terminology tends to be ambiguous, imprecise and unclear. Also, there is considerable room for errors and misunderstandings since people may have different views of the role of the software system. Furthermore, requirements vary and new requirements arise so that the specification is subject to frequent change. All these factors can lead to incomplete and inconsistent specifications that are difficult to validate against the requirements. People have advocated the use of formal specification languages to eliminate some of the problems associated with natural language, but formal languages have a grave disadvantage: they are not easily understood by untrained users.

Though it may seem that we are stuck between the over-flexibility of natural language and the potential incomprehensibility of formal languages, there is a solution. To improve the quality of specifications without loosing their readability we propose to restrict the use of natural language in specifications to a controlled subset with a well-defined syntax and semantics. On the one hand this subset should be expressive enough to allow natural usage by non-specialists, and on the other hand the language should be accurately and efficiently processable by a computer.

We suggest that the basic model of controlled natural language should cover the following constructs:

- simple declarative sentences of the form subject – predicate – object
- relative clauses, both subject and object modifying
- comparative constructions like *bigger than*, *smaller than* and *equal to*
- compound phrases like and-lists, or-lists, and-then-lists
- *if ... then* sentences
- negation like *does not, is not* and *has not*
- *yes/no* queries, *wh*-queries

Similar constructs have been proposed for the *computer-processable natural language* of Pulman and collaborators [Macias & Pulman 92, Pulman 94].

Furthermore, the controlled language is characterised by a vocabulary that comprises the usual closed word classes – prepositions, determiners etc. – and application-specific subsets of the open classes – e.g. nouns and verbs.

Users seem to be able to construct sentences in controlled natural language, and to avoid constructions that fall outside the bounds of the language, particularly when the system gives feedback of the analysed sentences in a paraphrased form [Epstein 85, Capindale & Crawford 89]. We are convinced that employing controlled natural language for specifications will be most successful when users are trained and willing to strive for clear writing.

An additional benefit of controlled natural language is that it may help finding an agreement concerning the correct interpretation of a specification. This is of utmost importance because a software specification will be read, interpreted, criticised, and rewritten, many times until a result is produced that is satisfactory to all participants.

## 3 Unification-Based Phrase Structure Grammars (PSGs)

The framework of phrase structure grammars builds the theoretical background for the syntactic and semantic processing of controlled language texts [Borsley 91]. For our implementation we are using Definite Clause Grammars enhanced by feature structures. These feature structures are written in GULP (Graph Unification Logic Programming) – a syntactic extension of Prolog that supports the implementation of unification-based PSGs by adding a notation for linearised feature structures [Covington 94], e.g.

```
case:nom .. agr:( person:third .. number:sg )
```

GULP adds to Prolog two operators and a number of system predicates. The first operator ':' binds a feature name to its value that can be a category. The second operator '..' joins one feature-value pair to the next.

GULP feature structures can be combined with the DCG formalism to yield a powerful lingua franca for natural language processing. Technically, this means introducing GULP feature structures as arguments into nodes of DCGs. Thus we can write

```
sentence -->  noun_phrase(case:nom .. agr:Person_Number),
              verb_phrase(agr:Person_Number).
```

The GULP translator accepts a Prolog program, scans it for linearised feature structures and converts them – by means of automatically built translation schemata – into an internal term representation called value list. Grammar rules are parsed top-down by the Prolog interpreter.

## 4 Discourse Representation Theory (DRT)

Correct understanding of a specification requires not only processing of individual sentences and their constituents, but also taking into account the way sentences are interrelated to express complex propositional structures. It has been recognised that aspects such as pronominal reference, tense and propositional attitudes cannot be successfully handled without taking the preceding discourse into consideration. We do this by employing discourse representation theory [Kamp & Reyle 93], and by extending our parser to extract the semantic structure of a sentence in the context of the preceding sentences.

DRT represents a multisentential natural language discourse in a single logical unit called a discourse representation structure (DRS). In general, a DRS $K$ is defined as an ordered pair $<U, Con>$ where $U$ is a set of discourse referents (discourse variables) and $Con$ is a set of conditions. The conditions $Con$ are either atomic (of the form $P(u_1, ..., u_n)$ or $u_1 = u_2$), or complex (negation, implication, or disjunction). A DRS is obtained through the application of a set of syntax-driven DRS construction rules $R$. These rules do not only examine the sentence under construction, but also the DRS that has been built so far. Thus we can define the meaning of a sentence $S$ as the function $M$ from a given DRS $K$ to an extended $K'$ induced by $R$.

**Simple DRSs**

The specification of the automated teller machine contains the two sentences

```
SimpleMat is a simple money dispenser.
It has a user interface.
```

Starting from the empty DRS $K_0$, the discourse representation structure for the two sentences is constructed by processing each sentence in turn. A first DRS $K_1$ corresponds to the processing of the first sentence. While the sentence is parsed top-down, the DRS $K_1$ is composed simultaneously, eventually yielding the following result.

```
X1 X2
-------------------------
named(X1,simplemat)
money_dispenser(X2)
simple(X2)
is(X1,X2)
```

The DRS $K_1$ says that the bearer of the name `SimpleMat` is identical with the object that was indicated by the noun phrase `a simple money dispenser`. This indefinite noun phrase contributes two conditions to the DRS: one condition for the compound noun `money dispenser` and one for the descriptive adjective `simple`. The function of the verb `to be` in the above discourse is to express that the two noun phrases have the same referent. To reflect this relation in the DRS a condition of the form `is(X1,X2)` has been introduced, which asserts that the objects represented by `X1` and `X2` coincide.

At this point, we will incorporate the second sentence into the established DRS $K_1$ by extending it to $K_2$. In order to do it, we have to find a suitable representation of the relation which holds between the anaphoric pronoun `it` and its antecedent in the first sentence. We will represent this information in the form of an equation, with the new referent on the left and the referent that is chosen as antecedent on the right of the equality sign. The referent chosen is the closest antecedent that agrees in case, number and gender.

Incorporating the whole information of the second sentence, we get three new conditions and obtain DRS $K_2$:

```
X1 X2 X3 X4
-------------------------
named(X1,simplemat)
money_dispenser(X2)
simple(X2)
is(X1,X2)
user_interface(X4)
have(X3,X4)
X3 = X1
```

**Complex DRSs**

DRSs that represent conditional, universal or negative sentences are complex, i.e. they contain sub-DRSs.

Sentences in which a subordinate *if*-clause combines with a main *then*-clause are usually referred to as conditional sentences. Such sentences serve the purpose of making hypothetical claims. The supposed *if*-clause is called the antecedent and the hypothetically asserted *then*-clause the consequent of the conditional. Intuitively, the consequent provides a situational description which extends that given by the antecedent.

In general, a conditional sentence of the form *if A then B* contributes to a DRS $K_0$ a condition of the form $K_1$ => $K_2$, where $K_1$ is a sub-DRS corresponding to $A$ and $K_2$ is the sub-DRS resulting from extending $K_1$ through the incorporation of $B$.

For instance, the conditional sentence of our specification

```
If the trap-door-algorithm calculates a number
then the number equals the check code.
```

is represented in DRT as:

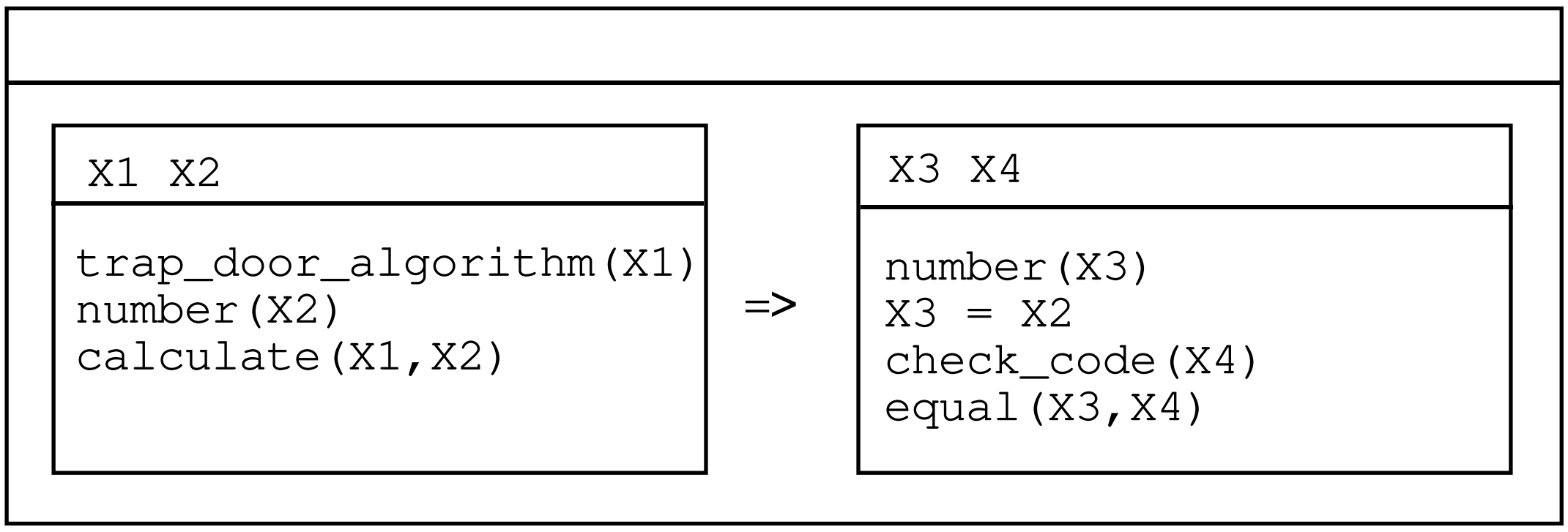

In terms of truth conditions, the above conditional $K_1$ => $K_2$ is satisfied if and only if there are individuals `X1` and `X2` that make the sub-DRS $K_1$ and the sub-DRS $K_2$ true simultaneously. This definition contrasts with classical logic where the implication is also true in the situation when the antecedent is false.

Note the unique reference use and the anaphoric use of the definite noun phrases. A unique reference `X1` of the definite noun phrase `the trap-door-algorithm` is created in the *if*-sub-DRS because no antecedent can be found in the superordinate DRS. The definite noun phrase `the number` is used anaphorically in the *then*-sub-DRS. Consequently, an equation of the form `X3 = X2` is generated, where `X2` is the discourse referent of the antecedent object noun phrase in the *if*-sub-DRS.

DRT claims that an anaphor can only refer to a discourse referent in the current DRS or in a DRS superordinate to it. Though this restriction makes correct predictions about the accessibility of antecedents to anaphors, it needs to be relaxed in practical applications to avoid contrived sentences.

As mentioned above DRSs are restricted formulas of predicate calculus, and resemble Horn clauses. All conditions in the antecedent are implicitly universally quantified and each condition in the consequent has an implicit existential quantifier contingent on the antecedent. The sub-DRS $K_1$ – on the left of the arrow – is called the restrictor of the quantifier, the one on the right – $K_2$ – its scope. In formalisms like predicate logic the semantic contributions of the words *if ... then* would have to be simulated by appropriate combinations of the universal quantifier and the implication connector. DRT seems to offer a much more natural representation for the systematic correlation between syntactic form and linguistic meaning of conditional sentences. This reflects the contextual role that a DRS was intended to play, namely mainly as a context for what is to be processed next, and not only as a representation of what has been processed already.

**DRT with Eventualities (DRT-E)**

DRT-E investigates further details of the semantics of verbs, taking into account the theory of underlying eventualities (events or states) and handling temporal and aspectual information [Kamp & Reyle 93, Reichenbach 47]. A prototypical implementation of DRT-E is described in [Brown 94].

Investigating sentences like

```
The customer enters the card.
SimpleMat checks the card.
```

and

```
Every customer has a personal code.
```

we recognise that the first two sentences are naturally understood as a report of temporally ordered events, while the second sentence describes something like a condition or state.

Verbs like `enter` or `check` introduce the existence of an event in much the same way as a noun phrase introduces the existence of an object. Events involve some kind of change in the universe of discourse, they persist through a certain interval of time and come eventually to a culmination point. They imply that some non-temporal condition, which is true when the event starts, is terminated by the event, and is replaced by further events.

States differ from events. A state verb such as `have` expresses a quality that is true indefinitely – it involves the continuation of a condition.

Sometimes the distinction between state-sentences and event-sentences is recognisable from the syntactic form of the verb, but it is well known that it is not the verb alone which decides about the eventuality introduced by a new sentence. The different thematic roles may exert a major influence.

In our approach, we represent the statement that `E1` is the event of `X1` entering `X2` as `enter(E1,X1,X2)` and use the special predicate `cul(E1,T1)` to express that the event `E1` culminates at time `T1`. The relation between the reference time `T1` of `E1` and the speech time `N` is established with the help of the additional predicate `at(T1,N)`. With these notational changes, the DRS looks like

```
N  E1  T1  X1  X2  E2  T2  X3  X4
---------------------------------
customer(X1)
card(X2)
enter(E1,X1,X2)
cul(E1,T1)
at(T1,N)

named(X3,simplemat)
card(X4)
check(E2,X3,X4)
X4 = X2
cul(E2,T2)
at(T2,N)
```

As mentioned above, a state is temporally extended and homogenous. It describes a static situation `S` that holds or does not hold at a given time `T`. The function of the verb `have` is twofold: it introduces a discourse referent `S` which represents a state of affair and it provides a descriptive characterisation of this state represented by `S`. We will retain this information as a predicate of the form `have(S,X1,X2)`. The additional predicate `hold(S,T)` asserts that `S` holds at `T` and the condition `at(T,N)` indicates that the eventuality described is located at the same time as the utterance time of the discourse that the DRS is taken to represent.

```
X1                      N  S  T  X2
-------------           -------------------
customer(X1)     =>     personal_code(X2)
                        have(S,X1,X2)
                        hold(S,T)
                        at(T,N)
```

**Ways to Investigate DRSs**

It is important to realise that a DRS can be investigated in several different ways.

First, it can be given a model-theoretic semantics by embedding it in a model. Second, a DRS can be manipulated deductively to infer further information using rules which operate only upon the structural content of the logical expressions. Third, a DRS can be investigated from a more psychological point of view as a contribution of building up a mental model of a language user.

The second and the third ways lead to the concept of knowledge assimilation [Kowalski 93]. According to this proof theoretic approach a DRS is processed by resource-constrained deduction and tested whether it can be added to a continuously changing theory. The terms truth and falsity of DRSs in model theory are replaced by the proof of consistency and inconsistency in the process of knowledge assimilation.

**From DRSs to Prolog**

Translating DRSs into Prolog clauses poses a problem – free variables in Prolog clauses have implicit universal quantifiers. It is not possible to translate the DRS for

```
SimpleMat is a simple money dispenser.
```

namely

| X1 X2 |
| --- |
| named(X1,simplemat)<br>money_dispenser(X2)<br>simple(X2)<br>is(X1,X2) |

into Prolog as

```
named(X1,simplemat).
money_dispenser(X2).
simple(X2).
is(X1,X2).
```

The first fact would mean "Anything is named SimpleMat". We would not even be able to say that the money dispenser is the same thing as the object named `SimpleMat`, because variables in different clauses are distinct even if they have the same name. What we need is a discourse marker for each existentially quantified entity. For that reason, a constant (integer) is randomly chosen to represent the individual `X1`. Then the DRS conditions – with the constants instantiated for the discourse referents – would be asserted in the knowledge base

```
named(1,simplemat).
money_dispenser(1).
simple(1).
```

Two additional problems arise when we translate conditions which use sub-DRSs. First, Prolog clauses cannot have two predicates in its consequent, i.e. clauses of the form

```
a,b :- c,d.
```

are not permitted. To deal with this problem, Covington and his collaborators introduce a special operator (`::-`) as intermediate representation for clauses with more than one consequent [Covington et al. 88]. Now we can write

```
a,b ::- c,d.
```

Since this rule cannot be asserted directly into the knowledge base it is split up into several Prolog clauses by distributing the consequents:

```
a :- c,d.
b :- c,d.
```

Second, if the consequent of a conditional sentence introduces new variables, these variables have implicit existential quantifiers which depend on the antecedent. Since Prolog cannot represent this dependence directly, we have to simulate it by a form of skolemisation as

```
card([2,X1]), have(X1,[2,X1]) ::- customer(X1).
```

respectively as the two Prolog clauses

```
card([2,X1])    :- customer(X1).
have(X1,[2,X1]) :- customer(X1).
```

The Prolog term `[2,X1]` can be interpreted as a value that is a function of the value of `X1`.

## 5 Overview of the Specification System

In this section we briefly describe the components of our specification system, most of which have already been implemented in a prototypical form.

The user enters specification text in controlled natural language which the *Dialog Component* forwards to the parser in tokenised form. Parsing errors and ambiguities to be resolved by the user are reported back by the dialog component. The user can also query the knowledge base in controlled natural language.

The *Parser* uses a predefined definite clause grammar with feature structures and a predefined linguistic lexicon to check sentences for syntactical correctness, and to generate syntax trees and sets of nested discourse representation structures.

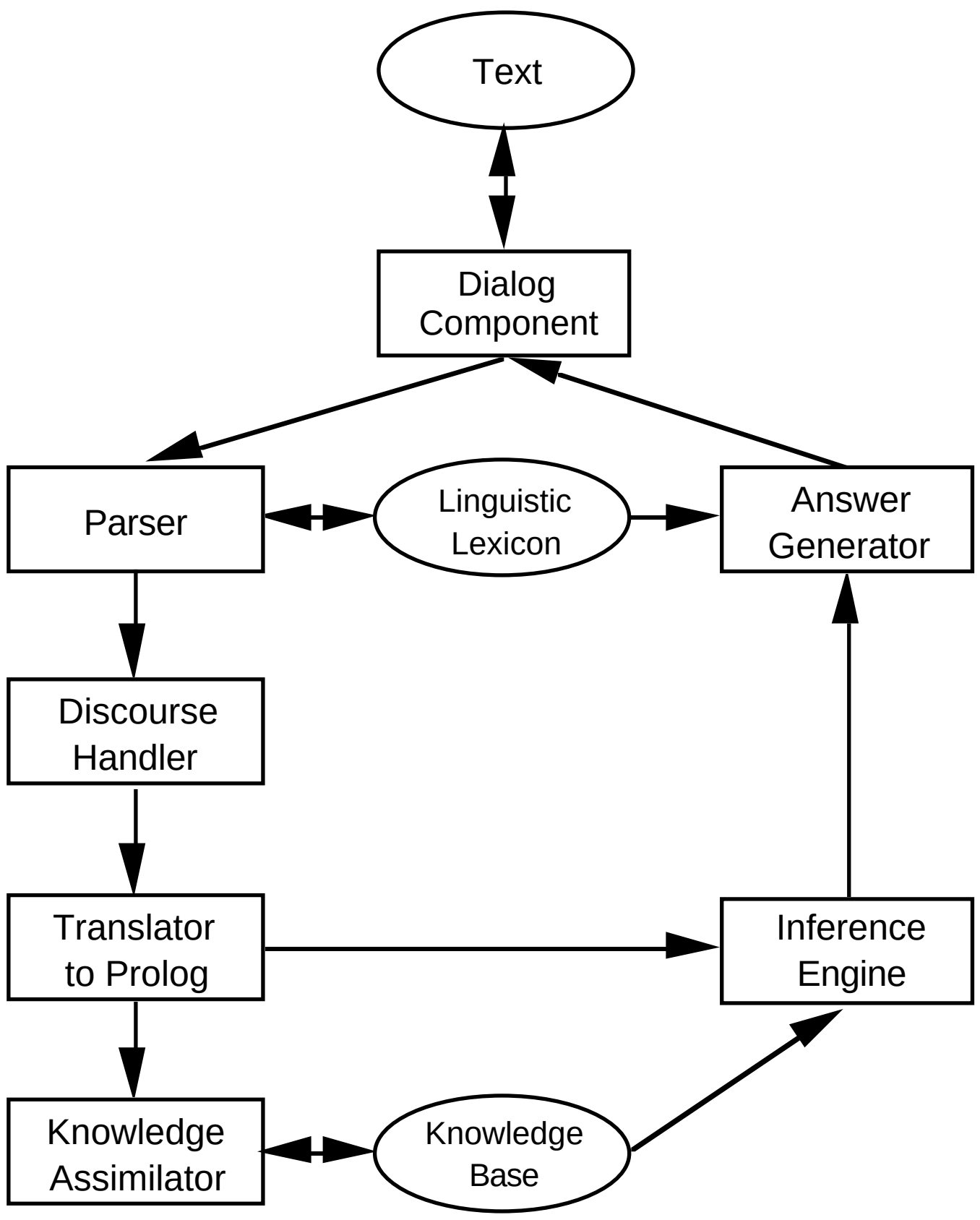

The *Linguistic Lexicon* contains an application-specific vocabulary. The lexicon can be modified by an editor invokable from the dialog component. This editor will be called automatically when the parser finds an undefined word.

The *Discourse Handler* analyses and resolves inter-text references and updates the discourse representation structures generated by the parser.

The *Translator* translates discourse representation structures into Prolog clauses. These Prolog clauses are either passed to the knowledge assimilator, or – in case of queries – to the inference engine.

The *Knowledge Assimilator* adds new knowledge to the knowledge base in a way that avoids inconsistency and redundancy.

The *Inference Engine* answers user queries with the help of the knowledge base. In a preliminary version the inference engine is just the Prolog interpreter.

The *Answer Generator* takes the answers of the inference engine, reformulates them in restricted natural language, and forwards them to the dialog component.

## 6 Using the Specification System

### An Example Translation

As a simple example we will demonstrate the translation of the sentence

```
Every customer has a card.
```

into Prolog. The discourse representation structure is built up by the parser and represented by a Prolog term of the form `drs(U,Con)`. `U` is a list of discourse referents represented by unique Prolog variables and `Con` is a list of terms containing these variables. For the example sentence the DRS becomes

```
drs([],
    [ ifthen(
    drs([X1],    [gender(X1,[m,f]), number(X1,sg), customer(X1)]),
    drs([X2,X1],[gender(X2,n), number(X2,sg), card(X2), have(X1,X2)]))
    ])
```

This initial DRS contains gender and number conditions that are employed for the resolution of anaphoric references. The discourse handler simplifies the DRS by eliminating the gender and number information, and by unifying terms. We get

```
drs([],
    [ ifthen(
    drs([X1],    [customer(X1)]),
    drs([X2,X1],[card(X2), have(X1,X2)]))
    ])
```

which is finally translated into the two Prolog clauses

```
card([2,X1])     :- customer(X1).
have(X1,[2,X1]) :- customer(X1).
```

that are added to the knowledge base. The Prolog term `[2,X1]` can be interpreted as a Skolem function assigning each customer his/her individual card.

### Querying the Knowledge Base

Like any other Prolog program the knowledge base can be queried by standard Prolog queries. But doing so would defy our tacit assumption that the user need not look at the internal representations of the specifications. Thus we allow certain classes of queries to be formulated in controlled natural language.

*Yes/No Queries*

The first class of queries just examines the factual information in the knowledge base. Let us assume that the user entered the text

```
SimpleMat is a simple money dispenser.
It has a user interface.
```

Now we can ask

```
Is SimpleMat a money dispenser?
```

This query will be translated into the Prolog query

```
named(1, simplemat), money_dispenser(1)
```

and the system will respond with

```
yes
```

Similarly

```
Does SimpleMat have a simple user interface?

no
```

*Wh-Queries*

Another class of queries contains pronouns like `who` and `what`. These pronouns are represented internally as variables that can be instantiated by Prolog terms during the inference process. The query

```
Who is a money dispenser?
```

is answered by the system in a form that shows the instantiation

```
[SimpleMat] is a money dispenser.
```

### Paraphrasing

To hide the internal representations of specifications, paraphrasing of analysed sentences in natural language is necessary in two situations – when the system gives feedback to the user, and when the user wants to examine the knowledge base.

*Feedback*

Let us assume that the user entered the text

```
SimpleMat is a simple money dispenser.
It has a user interface.
```

In this situation the system could reply like CLE [Alshawi 92] by

```
[SimpleMat] has a user interface.
```

to inform the user how the anaphoric reference was resolved. Similarly, for an ambiguous input

```
Every customer has a card.
```

the system could reply

```
Every customer has [an individual] card.
```

to tell the user which of the possible interpretations was chosen.

*Examining the Knowledge Base*

Prolog clauses of the knowledge base can be paraphrased in a simple way with the help of predefined schemata that access the linguistic lexicon. Here is an example schema.

```
named(X, <proper noun>)
<adjective>(Y)
<noun>(Y)
is(X,Y)
```

`<proper noun>`$^g$ `is a/an <adjective>`$^g$ `<noun>`$^g$`.`

Terms in angular brackets are schema variables that denote the relation names of preterminals. In the paraphrased sentence schema variables are replaced by the pertinent graphemes as indicated by the g superscript. With the help of this schema the Prolog clauses

```
named(1, john).
known(2).
customer(2).
is(1,2).
```

can be paraphrased as

```
John is a known customer.
```

Since the specification language is controlled and the translation into Prolog is very regular, paraphrasing schemata can be readily derived though they may not be as simple as in the example. Paraphrasing schemata can either be represented as Prolog terms or as simple grammars.

**Executing the Specification**

In the preceding discussion we have regarded the knowledge base only as a data base to be queried or examined. However, the knowledge base can also be used for simulation or prototyping by executing it. In our example specification, this means executing/running the specification of the automated teller machine. As it stands the specification does not provide all the necessary information and needs to be enhanced in three ways.

- Most importantly, an order of events has to be established, e.g. we have to make sure that during the simulation the event of entering a card has to precede the event of checking it. [Ishihara et al. 92] who translate natural language specifications into algebraic ones use contextual dependency and properties of data types to establish the correct order of events. In our approach based on discourse representation theory the order of events is to a great extent established when we introduce eventualities (events and states) into the processing of our controlled natural language. Following Kamp we interpret the sentences

  ```
  The customer enters the card.
  SimpleMat checks the card.
  ```

as an *entering event* temporally followed by a *checking event*, and the sentences

```
The customer enters the card.
SimpleMat is checking the card.
```

as an *entering event* that temporally overlaps with a *checking state*. This leads to an ordering of events or times, respectively. On the basis of this information a simple forward chaining meta-interpreter can execute the Prolog clauses in their correct order.

- Many relations representing events are not only truth-functional, but also cause side-effects, e.g. I/O operations. The required side-effects can be defined by interface predicates, e.g.

  ```
  enter(X, Y) :- prompt_read(['Enter your card'], Y).
  ```

  that depend on the simulation environment. One could, for example, envisage that the interface predicates do not simply simulate the automated teller machine but cause the execution of a real automated teller machine.

- Finally, the execution needs some situation specific information. For example, to execute the Prolog clauses

  ```
  card([2,X1])       :- customer(X1).
  have(X1,[2,X1]) :- customer(X1).
  ```

  derived from the input

  ```
  Every customer has a card.
  ```

  the goal `customer(X1)` must be provable. We can either provide the relevant Prolog facts, or more conveniently, get the information by querying the user.

## 7 Conclusion and Future Research

The present prototypical implementation of our system proves that controlled natural language can be used for the non-trivial specification of an automated teller machine, and that the specification can be translated as coherent text into Prolog clauses.

More work needs to be done, however, to turn the prototype into a useful tool. Besides extending the functionality of the existing system, we plan to enhance it in at least two ways. To specify time-dependencies explicitly, we will add constructs like *before*, *after* and *when* to the controlled natural language, and introduce event semantics into the DRSs. Though natural language – even in a controlled form – is a universal specification notation, we believe that it is not always the optimal one. Thus we will add graphical notations, e.g. for window-oriented user interfaces, and specific notations for algorithms.

## Acknowledgements

We would like to thank Jaume Agusti, Michael Hess, David Robertson, Wamberto Vasconcelos and Martin Volk for many stimulating discussions, and the anonymous reviewers of the extended abstract of this paper for valuable advice.